\documentclass[runningheads]{llncs}
\usepackage{graphicx}

\usepackage{tikz}
\usepackage{comment}
\usepackage{amsmath,amssymb} 
\usepackage{color}
\usepackage{booktabs}
\usepackage{multirow}
\usepackage{colortbl}
\usepackage{caption}  
\usepackage{wrapfig}
\usepackage{hhline}
\usepackage{caption}
\usepackage{hyperref}
\usepackage[accsupp]{axessibility}  

\begin{document}
\pagestyle{headings}
\mainmatter
\def\ECCVSubNumber{205}  

\title{Contrast-Phys: Unsupervised Video-based Remote Physiological Measurement via Spatiotemporal Contrast} 

\titlerunning{Contrast-Phys}
%
\author{Zhaodong Sun \and Xiaobai Li}
\authorrunning{Z. Sun and X. Li}
%
\institute{Center for Machine Vision and Signal Analysis, University of Oulu, Oulu, Finland
\email{\{zhaodong.sun, xiaobai.li\}@oulu.fi}}
\maketitle

\begin{abstract}
Video-based remote physiological measurement utilizes face videos to measure the blood volume change signal, which is also called remote photoplethysmography (rPPG). Supervised methods for rPPG measurements achieve state-of-the-art performance. However, supervised rPPG methods require face videos and ground truth physiological signals for model training. In this paper, we propose an unsupervised rPPG measurement method that does not require ground truth signals for training. We use a 3DCNN model to generate multiple rPPG signals from each video in different spatiotemporal locations and train the model with a contrastive loss where rPPG signals from the same video are pulled together while those from different videos are pushed away. We test on five public datasets, including RGB videos and NIR videos. The results show that our method outperforms the previous unsupervised baseline and achieves accuracies very close to the current best supervised rPPG methods on all five datasets. Furthermore, we also demonstrate that our approach can run at a much faster speed and is more robust to noises than the previous unsupervised baseline. Our code is available at \url{https://github.com/zhaodongsun/contrast-phys}.

\end{abstract}

\section{Introduction}

Traditional physiological measurement requires skin-contact sensors to measure physiological signals such as contact photoplethysmography (PPG) and electrocardiography (ECG). Some physiological parameters like heart rate (HR), respiration frequency (RF), and heart rate variability (HRV) can be derived from PPG signals for healthcare \cite{shi2019atrial,yan2018contact} and emotion analysis \cite{yu2021facial,mcduff2014remote,sabour2021ubfc}. However, the skin-contact physiological measurement requires specific biomedical equipment like pulse oximeters, and contact sensors may cause discomfort and skin irritation. Remote physiological measurement uses a camera to record face videos for measuring remote photoplethysmography (rPPG). The weak color change in faces can be captured by cameras to obtain the rPPG signal from which several physiological parameters such as HR, RF, and HRV \cite{poh2010advancements} can be measured. Video-based physiological measurement only requires off-the-shelf cameras rather than professional biomedical sensors, and is not constrained by physical distance, which has a great potential for remote healthcare \cite{shi2019atrial,yan2018contact} and emotion analysis applications \cite{yu2021facial,mcduff2014remote,sabour2021ubfc}.

In earlier rPPG studies \cite{verkruysse2008remote,poh2010advancements,de2013robust,wang2016algorithmic}, researchers proposed handcrafted features to extract rPPG signals. Later, some deep learning (DL)-based methods \cite{chen2018deepphys,vspetlik2018visual,yu2019remoteBMVC,yu2019remote,lee2020meta,NEURIPS2020_e1228be4,niu2019rhythmnet,niu2020video,lu2021dual,nowara2021benefit} were proposed, which employed supervised approaches with various network architectures for measuring rPPG signals. On one side, under certain circumstances, e.g., when head motions are involved or the videos are heterogeneous, DL-based methods could be more robust than the traditional handcrafted approaches. On the other side, DL-based rPPG methods require a large-scale dataset including face videos and ground truth physiological signals. Although face videos are comparatively easy to obtain in large amount, it is expensive to get the ground truth physiological signals which are measured by contact sensors, and synchronized with the face videos.

Can we only use face videos without ground truth physiological signals to train models for rPPG measurement? Gideon and Stent \cite{gideon2021way} proposed a self-supervised method to train rPPG measurement models without labels. They first downsample a video to get the downsampled rPPG (negative sample) and upsample the downsampled rPPG to get the reconstructed rPPG (positive sample). They also used the original video to get the anchor rPPG. Then a triplet loss is used to pull together the positive and anchor samples, and push away negative and anchor samples. However, their method has three problems. 1) They have to forward one video into their backbone model twice, which causes an extra computation burden. 2) There is still a significant gap between the performance of their unsupervised method and the state-of-the-art supervised rPPG methods \cite{vspetlik2018visual,yu2019remoteBMVC,nowara2020near}. 3) They showed in their paper \cite{gideon2021way} that their method is easily impacted by external periodic noise.

We propose a new unsupervised method (Contrast-Phys) to tackle the problems above. Our method was built on four observations about rPPG. 1) \textbf{rPPG spatial similarity}: rPPG signals measured from different facial areas have similar power spectrum densities (PSDs) 2) \textbf{rPPG temporal similarity}: In a short time, two rPPG signals (e.g., two consecutive 5s clips) usually present similar PSDs as the HR tends to take smooth transits in most cases. 3) \textbf{Cross-video rPPG dissimilarity}: The PSDs of rPPG signals from different videos are different. 4) \textbf{HR range constraint}: The HR should fall between 40 and 250 beats per minute (bpm), so we only care about the PSD in the frequency interval between 0.66 Hz and 4.16 Hz.

We propose to use a 3D convolutional neural network (3DCNN) to process an input video to get a spatiotemporal rPPG (ST-rPPG) block. The ST-rPPG block contains multiple rPPG signals along three dimensions of height, width, and time. According to the rPPG spatiotemporal similarity, we can randomly sample rPPG signals from the same video in different spatiotemporal locations and pull them together. According to the cross-video rPPG dissimilarity, the sampled rPPG signals from different videos are pushed away. The whole procedures are shown in Fig. \ref{fig:diagram} and Fig. \ref{fig:sampler}.

The contributions of this work are 1) We propose a novel rPPG representation called spatiotemporal rPPG (ST-rPPG) block to obtain rPPG signals in spatiotemporal dimensions. 2) Based on four observations about rPPG, including rPPG spatiotemporal similarity and cross-video rPPG dissimilarity, we propose an unsupervised method based on contrastive learning. 3) We conduct experiments on five rPPG datasets (PURE \cite{stricker2014non}, UBFC-rPPG \cite{bobbia2019unsupervised}, OBF \cite{li2018obf}, MR-NIRP \cite{nowara2018sparseppg}, and MMSE-HR \cite{zhang2016multimodal}) including RGB and NIR videos under various scenarios. Our method outperforms the previous unsupervised baseline \cite{gideon2021way} and achieves very close performance to supervised rPPG methods. Contrast-Phys also shows significant advantages with fast running speed and noise robustness compared to the previous unsupervised baseline.

\section{Related Work}

\subsubsection{Video-Based Remote Physiological Measurement.}
Verkruysse et al. \cite{verkruysse2008remote} first proposed that rPPG can be measured from face videos from the green channel. Several traditional handcraft methods \cite{poh2010advancements,de2013robust,de2014improved,wang2016algorithmic,lam2015robust,tulyakov2016self,wang2014exploiting} were proposed to further improve rPPG signal quality. Most rPPG methods proposed in earlier years used handcrafted procedures and did not need datasets for training, which are referred to as traditional methods. Deep learning (DL) methods for rPPG measurement are rapidly emerging. Several studies \cite{chen2018deepphys,vspetlik2018visual,NEURIPS2020_e1228be4,nowara2021benefit} used a 2D convolutional neural network (2DCNN) with two consecutive video frames as the input for rPPG measurement. Another type of DL-based methods \cite{niu2019rhythmnet,niu2020video,lu2021dual} used a spatial-temporal signal map extracted from different facial areas as the input to feed into a 2DCNN model. Recently, 3DCNN-based methods \cite{yu2019remote,yu2019remoteBMVC,gideon2021way} were proposed to achieved good performance on compressed videos \cite{yu2019remote}. The DL-based methods require both face videos and ground truth physiological signals, so we refer to them as supervised methods. Recently, Gideon and Stent \cite{gideon2021way} proposed an unsupervised method to train a DL model without ground truth physiological signals. However, their method falls behind some supervised methods and is not robust to external noise. In addition, the running speed is not satisfactory.

\subsubsection{Contrastive Learning.}
Contrastive learning is a self-supervised learning method widely used in video and image feature embedding, which facilitates downstream task training and small dataset fine-tuning \cite{hadsell2006dimensionality,schroff2015facenet,van2018representation,chen2020simple,tian2020contrastive,he2020momentum,grill2020bootstrap,misra2020self,qian2021spatiotemporal}. A DL model working as a feature extractor maps a high-dimensional image/video into a low dimensional feature vector. To train this DL feature extractor, features from different views of the same sample (positive pairs) are pulled together, while features from views of different samples (negative pairs) are pushed away. Data augmentations (such as cropping, blurring \cite{chen2020simple}, and temporal sampling \cite{qian2021spatiotemporal}) are used to obtain different views of the same sample so that the learned features are invariant to some augmentations. Previous works mentioned above use contrastive learning to let a DL model produce abstract features for downstream tasks such as image classification \cite{chen2020simple}, video classification \cite{qian2021spatiotemporal}, face recognition \cite{schroff2015facenet}. On the other hand, our work uses contrastive learning to directly let a DL model produce rPPG signals, enabling unsupervised learning without ground truth physiological signals.


\section{Observations about rPPG}\label{sec:observation}
This section describes four observations about rPPG, which are the precondition to design our method and enable unsupervised learning.


\subsubsection{rPPG Spatial Similarity.} 
\begin{figure}[hbt!]
\centering
\begin{minipage}[b]{\linewidth}
  \centering
  \centerline{\includegraphics[width=\linewidth]{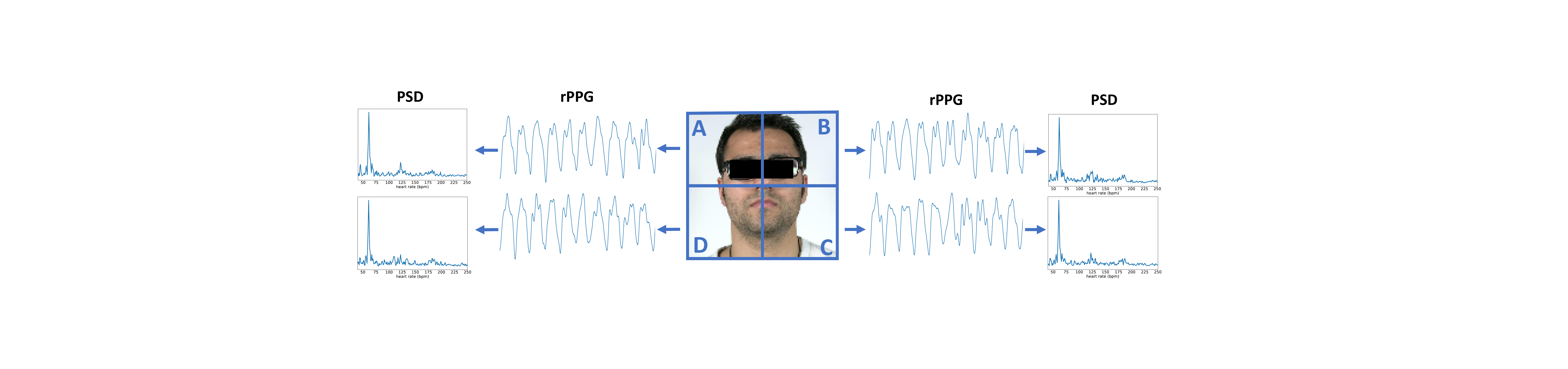}}
\end{minipage}
\caption{Illustration of rPPG spatial similarity. The rPPG signals from four facial areas (A, B, C, D) have similar waveforms and power spectrum densities (PSDs)}
\label{fig:rppg_spatial}
\end{figure}

rPPG signals from different facial areas have similar waveforms, and their PSDs are also similar. Several works \cite{lam2015robust,tulyakov2016self,kumar2015distanceppg,wang2014exploiting,wang2019discriminative,liu20163d,liu2018remote} also exploited rPPG spatial similarity to design their methods. There might be small phase and amplitude differences between two rPPG signals from two different body skin areas \cite{kamshilin2011photoplethysmographic,kamshilin2013variability}. However, when rPPG waveforms are transformed to PSDs, the phase information is erased, and the amplitude can be normalized to cancel the amplitude difference. In Fig. \ref{fig:rppg_spatial}, the rPPG waveforms from four spatial areas are similar, and they have the same peaks in PSDs.

\subsubsection{rPPG Temporal Similarity.}
\begin{figure}[hbt!]
\centering
\begin{minipage}[b]{\linewidth}
  \centering
  \centerline{\includegraphics[width=\linewidth]{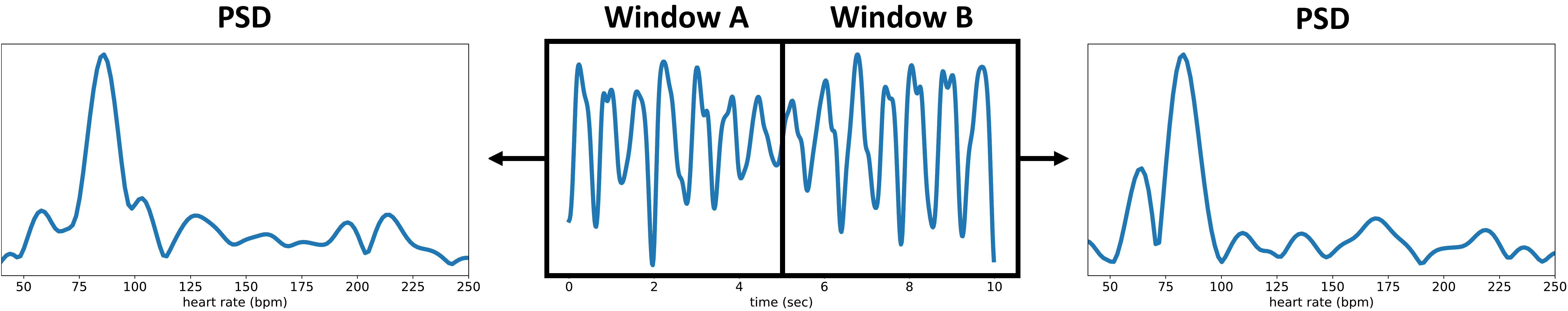}}
\end{minipage}
\caption{Illustration of rPPG temporal similarity. The rPPG signals from two temporal windows (A, B) have similar PSDs}
\label{fig:rppg_temporal}
\end{figure}

The HR does not change rapidly in a short term \cite{gideon2021way}. Stricker et al. \cite{stricker2014non} also found that the HR varies slightly in a short time interval in their dataset. Since the HR has a dominant peak in PSD, the PSD does not change rapidly, either. If we randomly sample several small windows from a short rPPG clip (e.g., 10s), the PSDs of these windows should be similar. In Fig. \ref{fig:rppg_temporal}, we sample two 5s windows from a short 10s rPPG signal and get the PSDs of these two windows. The two PSDs are similar and have sharp peaks at the same frequency. Since this observation is only valid under the condition of short-term rPPG signals, in the following sensitivity analysis part, we will discuss the influence of the signal length on our model performance. Overall, we can use the equation $\text{PSD}\big\{ G\big(v(t_1 \to t_1 + \Delta t, \mathcal{H}_1, \mathcal{W}_1)\big) \big\}
\approx \text{PSD}\big\{ G\big(v(t_2 \to t_2 + \Delta t, \mathcal{H}_2, \mathcal{W}_2)\big) \big\}$ to describe spatiotemporal rPPG similarity. $v \in \mathbb{R}^{T \times H \times W \times 3}$ is a facial video, $G$ is an rPPG measurement algorithm. We can choose one facial area with a set of height $\mathcal{H}_1$ and width $\mathcal{W}_1$, and a time interval $t_1 \to t_1+\Delta t$ from video $v$ to achieve one rPPG signal. We can achieve another rPPG signal similarly from the same video with $\mathcal{H}_2$, $\mathcal{W}_2$, and $t_2 \to t_2+\Delta t$. $|t_1 - t_2|$ should be small to satisfy the condition of short-term rPPG signals.

\subsubsection{Cross-video rPPG Dissimilarity.}

\begin{wrapfigure}{r}{0.5\linewidth}
\centering
\begin{minipage}[b]{\linewidth}
  \centering
  \centerline{\includegraphics[width=\linewidth]{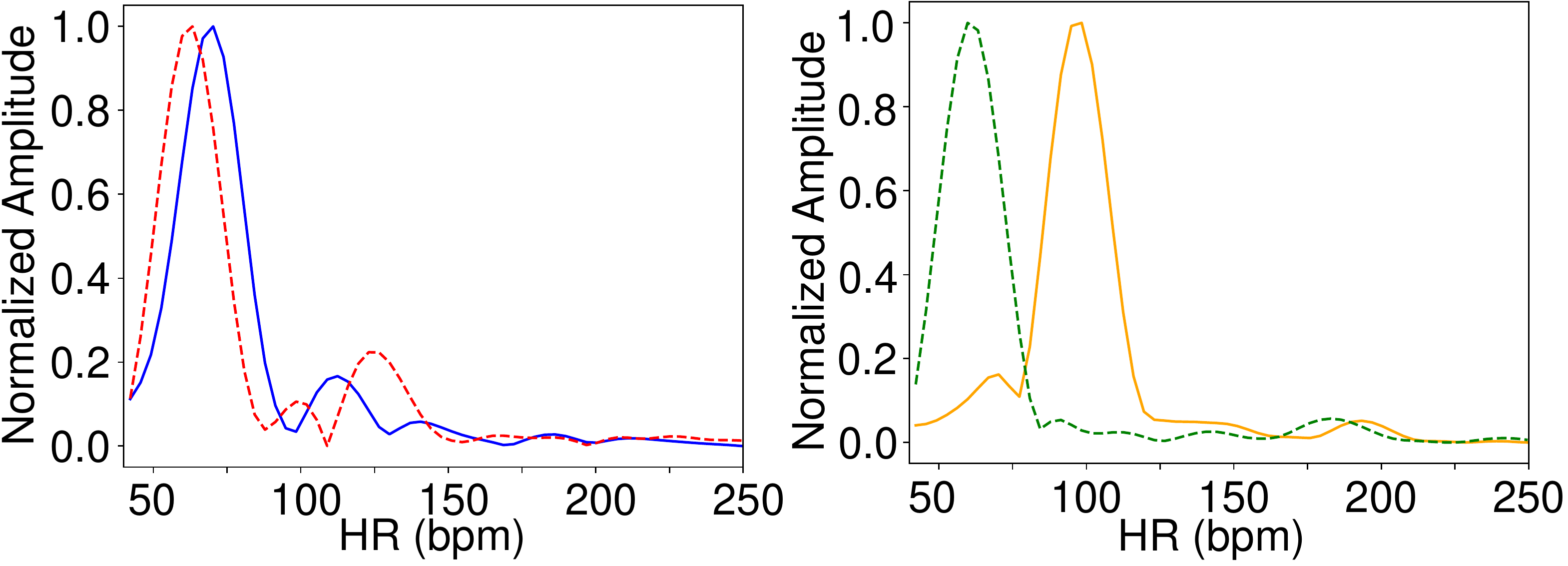}}
\end{minipage}
\caption{The most similar (left) and most different (right) cross-video PSD pairs in the OBF dataset.}
\label{fig:cross-video}
\end{wrapfigure}

We assume rPPG signals from different face videos have different PSDs. Each video is recorded with different people and different physiological states (such as exercises and emotion status), so the HRs across different videos are likely different \cite{hr_factor}. Even though the HRs might be similar between two videos, the PSDs might still be different since PSD also contains other physiological factors such as respiration rate \cite{chen2020modulation} and HRV \cite{pai2021hrvcam} which are unlikely to be all the same between two videos. To further validate the observation, we calculate the mean squared error for all cross-video PSD pairs in the OBF dataset \cite{li2018obf} and show the most similar and most different cross-video PSD pairs in Fig. \ref{fig:cross-video}. It can be observed that the main cross-video PSD difference is the heart rate peak. The following equation describes cross-video rPPG dissimilarity. $\text{PSD}\big\{ G\big(v(t_1 \to t_1 + \Delta t, \mathcal{H}_1, \mathcal{W}_1)\big) \big\}
\neq \text{PSD}\big\{ G\big(v^{\prime}(t_2 \to t_2 + \Delta t, \mathcal{H}_2, \mathcal{W}_2)\big) \big\}$ where $v$ and $v^{\prime}$ are two different videos. We can choose facial areas and time intervals from these two videos. The PSDs of the two rPPG signals should be different.


\subsubsection{HR Range Constraint.}
The HR range for most people is between 40 and 250 bpm \cite{hr_interval}. Most works \cite{poh2010advancements,li2014remote} use this HR range for rPPG signal filtering and find the highest peak to estimate the HR. Therefore, our method will focus on PSD between 0.66 Hz and 4.16 Hz.

\section{Method}

\begin{figure}[hbt!]
\centering
\begin{minipage}[b]{\linewidth}
  \centering
  \centerline{\includegraphics[width=\linewidth]{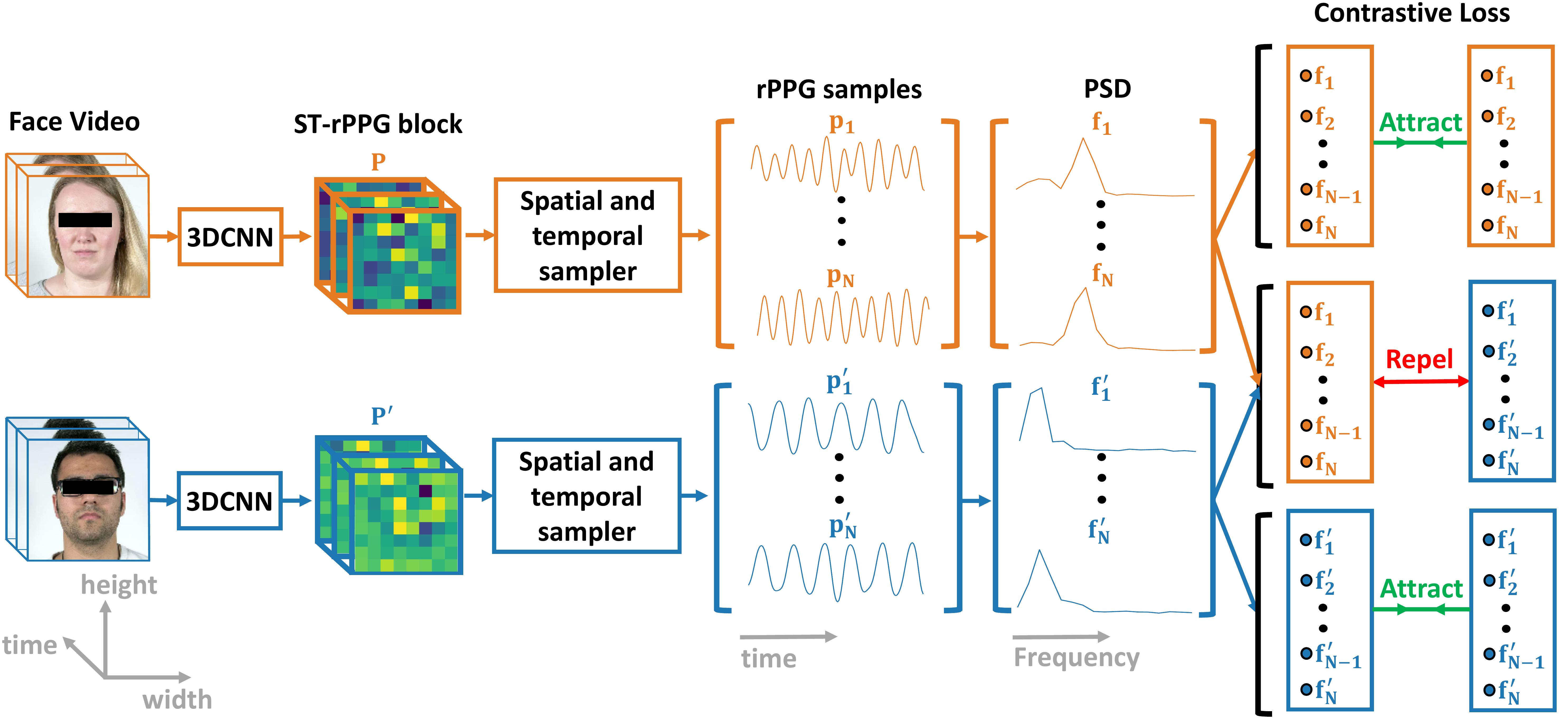}}
\end{minipage}
\caption{Contrast-Phys Diagram. A pair of videos are fed into the same 3DCNN to generate a pair of ST-rPPG blocks. Multiple rPPG samples are sampled from the ST-rPPG blocks (The spatiotemporal sampler is illustrated in Fig. \ref{fig:sampler}) and converted to PSDs. The PSDs from the same video are attracted while the PSDs from different videos are repelled.}
\label{fig:diagram}
\end{figure}

The overview of Contrast-Phys is shown in Fig. \ref{fig:diagram} and Fig. \ref{fig:sampler}. We describe the procedures of Contrast-Phys in this section.

\subsection{Preprocessing}\label{sec:preprocessing}
The original videos are firstly preprocessed to crop the face to get the face video shown in Fig. \ref{fig:diagram} left. Facial landmarks are generated using OpenFace \cite{baltrusaitis2018openface}. We first get the minimum and maximum horizontal and vertical coordinates of the landmarks to locate the central facial point for each frame. The bounding box size is 1.2 times the vertical coordinate range of landmarks from the first frame and is fixed for the following frames. After getting the central facial point of each frame and the size of the bounding box, we crop the face from each frame. The cropped faces are resized to $128 \times 128$, which are ready to be fed into our model.

\subsection{Spatiotemporal rPPG (ST-rPPG) Block Representation}\label{sec:ST-rPPG block representation}

We modify 3DCNN-based PhysNet \cite{yu2019remoteBMVC} to get the ST-rPPG block representation. The modified model has an input RGB video with the shape of $T \times 128 \times 128 \times 3$ where $T$ is the number of frames. In the last stage of our model, we use adaptive average pooling to downsample along spatial dimensions, which can control the output spatial dimension length. This modification allows our model to output a spatiotemporal rPPG block with the shape of $T \times S \times S$ where $S$ is spatial dimension length as shown in Fig. \ref{fig:sampler}. More details about the 3DCNN model are described in the supplementary material.

The ST-rPPG block is a collection of rPPG signals in spatiotemporal dimensions. We use $P \in \mathbb{R}^{T \times S \times S}$ to denote the ST-rPPG block. Suppose we choose a spatial location $(h, w)$ in the ST-rPPG block. In that case, the corresponding rPPG signal in this position is $P(\cdot, h, w)$ which is extracted from the receptive field of this spatial position in the original video. We can deduce that when the spatial dimension length $S$ is small, each spatial position in the ST-rPPG block has a larger receptive field. The receptive field of each spatial position in the ST-rPPG block can cover part of the facial region, which means all spatial positions in the ST-rPPG block can include rPPG information.

\subsection{rPPG Spatiotemporal Sampling}\label{sec:rppg sampling}

\begin{figure}[hbt!]
\centering
\begin{minipage}[b]{\linewidth}
  \centering
  \centerline{\includegraphics[width=\linewidth]{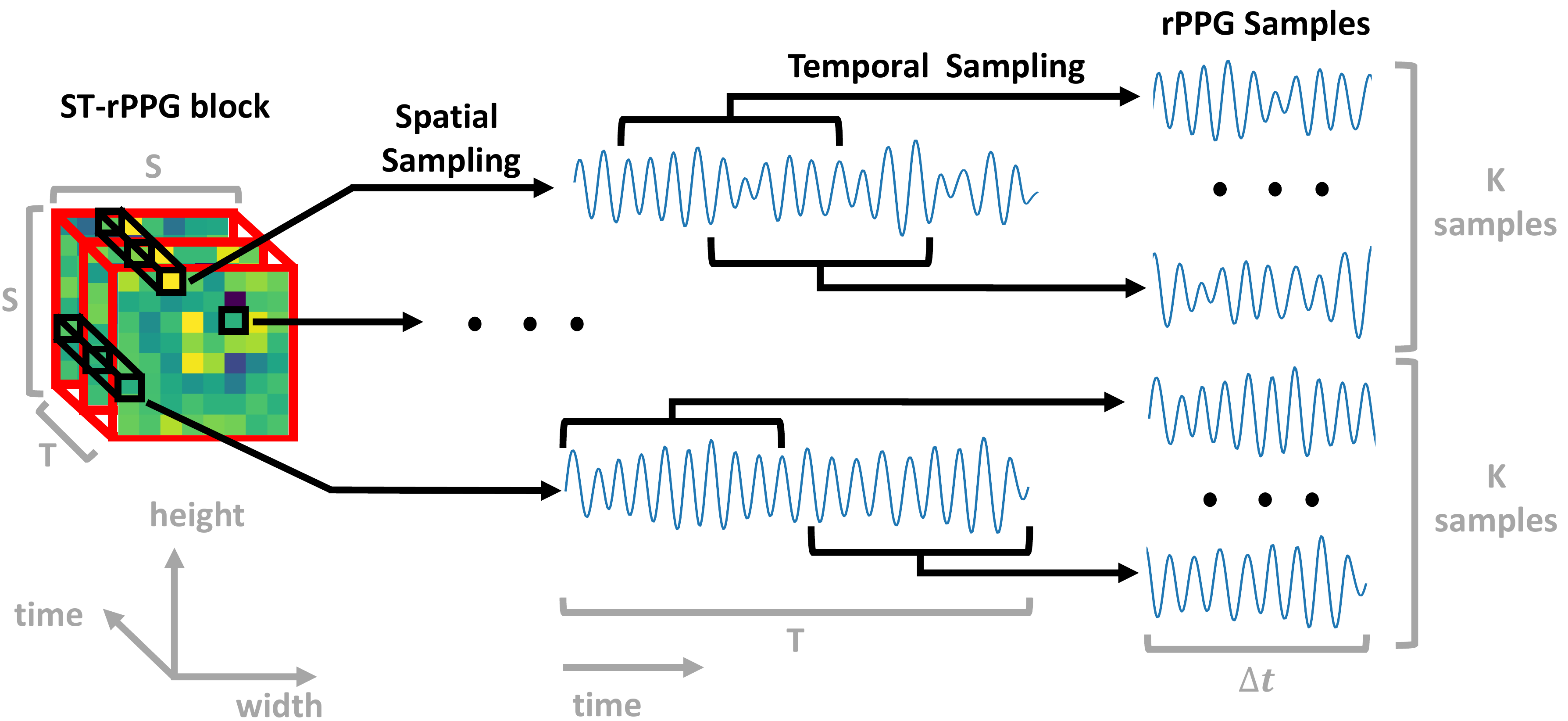}}
\end{minipage}
\caption{Spatiotemporal Sampler}
\label{fig:sampler}
\end{figure}

Several rPPG signals are sampled from the ST-rPPG block as illustrated in Fig. \ref{fig:sampler}. For spatial sampling, we can get the rPPG signal $P(\cdot, h, w)$ at one spatial position. For temporal sampling, we can sample a short time interval from $P(\cdot, h, w)$, and the final spatiotemporal sample is $P(t \to t+\Delta t, h, w)$ where $h$ and $w$ are the spatial position, $t$ is the starting time, and $\Delta t$ is the time interval length. For one ST-rPPG block, we will loop over all spatial positions and sample $K$ rPPG clips with a randomly chosen starting time $t$ for each spatial position. Therefore, we can get $S \cdot S \cdot K$ rPPG clips from the ST-rPPG block. The more detailed sampling procedures are in our supplementary material. The sampling procedures above are used during model training. After our model is trained and used for testing, we can directly average ST-rPPG over spatial dimensions to get the rPPG signal.

\subsection{Contrastive Loss Function}\label{sec:contrastive loss function}
As illustrated in Fig. \ref{fig:diagram}, we have two different videos randomly chosen from a dataset as the input. For one video, we can get one ST-rPPG block $P$, a set of rPPG samples $[p_1, \dots, p_N]$ and the corresponding PSDs $[f_1, \dots, f_N]$. For another video, we can get one ST-rPPG block $P^{\prime}$, one set of rPPG samples $[p^{\prime}_1, \dots, p^{\prime}_N]$ and the corresponding PSDs $[f^{\prime}_1, \dots, f^{\prime}_N]$ in the same way. As shown in Fig. \ref{fig:diagram} right, the principle of our contrastive loss is to pull together PSDs from the same video and push away PSDs from different videos. It is noted that we only use the PSD between 0.66 Hz and 4.16 Hz according to the HR range constraint in Sec. \ref{sec:observation}. 

\subsubsection{Positive Loss Term.}
According to rPPG spatiotemporal similarity, we can conclude that the PSDs from the spatiotemporal sampling of the same ST-rPPG block should be similar. We can use the following equations to describe this property for the two input videos. For one video, $\text{PSD}\big\{P(t_1 \to t_1+\Delta t, h_1, w_1)\big\} \approx \text{PSD}\big\{P(t_2 \to t_2+\Delta t, h_2, w_2)\big\} \implies f_i \approx f_j, i \neq j$. For another video, $\text{PSD}\big\{P^{\prime}(t_1 \to t_1+\Delta t, h_1, w_1)\big\} \approx \text{PSD}\big\{P^{\prime}(t_2 \to t_2+\Delta t, h_2, w_2)\big\} \implies f^{\prime}_i \approx f^{\prime}_j, i \neq j$.


We can use the mean squared error as the loss function to pull together PSDs (positive pairs) from the same video. The positive loss term $L_p$ is shown below, which is normalized with the total number of positive pairs.

\begin{equation}
L_p =  \sum_{i=1}^{N} \sum_{\substack{j=1 \\j \neq i}}^{N} \big(\parallel f_i - f_j \parallel^2 + \parallel f^{\prime}_i - f^{\prime}_j \parallel^2 \big) / \big(2N(N-1)\big)
\end{equation}

\subsubsection{Negative Loss Term.}
According to the cross-video rPPG dissimilarity, we can conclude that the PSDs from the spatiotemporal sampling of the two different ST-rPPG blocks should be different. We can use the following equation to describe this property for the two input videos. $\text{PSD}\big\{P(t_1 \to t_1+\Delta t, h_1, w_1)\big\}
\neq \text{PSD}\big\{P^{\prime}(t_2 \to t_2+\Delta t, h_2, w_2)\big\} \implies f_i \neq f^{\prime}_j$


We use the negative mean squared error as the loss function to push away PSDs (negative pairs) from two different videos. The negative loss term $L_n$ is shown below, which is normalized with the total number of negative pairs.

\begin{equation}
L_n = - \sum_{i=1}^{N} \sum_{j=1}^{N} \parallel f_i - f^{\prime}_j \parallel^2  / N^2
\end{equation}

The overall loss function is $L  = L_p + L_n$, which is the sum of the positive and negative loss terms.


\subsubsection{Why Our Method Works.}
Our four rPPG observations are constraints to make the model learn to keep rPPG and exclude noises since noises do not satisfy the observations. Noises that appear in a small local region such as periodical eye blinking are excluded since the noises violate rPPG spatial similarity. Noises such as head motions/facial expressions that do not have a temporal constant frequency are excluded since they violate rPPG temporal similarity. Noises such as light flickering that exceed the heart rate range are also excluded due to the heart rate range constraint. Cross-video dissimilarity in the loss can make two videos' PSDs discriminative and show clear heart rate peaks since heart rate peaks are one of the discriminative clues between two videos' PSDs.

\section{Experiments}

\subsection{Experimental Setup and Metrics}

\subsubsection{Datasets.}
We test five commonly used rPPG datasets covering RGB and NIR videos recorded under various scenarios. PURE \cite{stricker2014non}, UBFC-rPPG \cite{bobbia2019unsupervised}, OBF \cite{li2018obf} and MR-NIRP \cite{nowara2018sparseppg} are used for the intra-dataset testing. MMSE-HR \cite{zhang2016multimodal} is used for cross-dataset testing. \textbf{PURE} has ten subjects' face videos recorded in six different setups, including steady and motion tasks. We use the same experimental protocol as in \cite{vspetlik2018visual,lu2021dual} to divide the training and test set. \textbf{UBFC-rPPG} includes facial videos from 42 subjects who were playing a mathematical game to increase their HRs. We use the same protocol as in \cite{lu2021dual} for the train-test split and evaluation. \textbf{OBF} has 100 healthy subjects' videos recorded before and after exercises. We use subject-independent ten-fold cross-validation as used in \cite{yu2019remoteBMVC,yu2019remote,niu2020video} to make fair comparison with previous results. \textbf{MR-NIRP} has NIR videos from eight subjects sitting still or doing motion tasks. The dataset is challenging due to its small scale, and weak rPPG signals in NIR \cite{martinez2011optimal,vizbara2013comparison}. We will use a leave-one-subject-out cross-validation protocol for our experiments. \textbf{MMSE-HR} has 102 videos from 40 subjects recorded in emotion elicitation experiments. This dataset is also challenging since spontaneous facial expressions, and head motions are involved. More details about these datasets can be found in the supplementary material.

\subsubsection{Experimental Setup.} 
This part shows our experimental setup during training and testing. For the spatiotemporal sampler, we evaluate different spatial resolutions and time lengths of ST-rPPG blocks in the sensitivity analysis part. According to the results, we fix the parameters in the other experiments as follows. We set $K=4$, which means, for each spatial position in the ST-rPPG block, four rPPG samples are randomly chosen. We set the spatial resolution of the ST-rPPG block as $2 \times 2$, and the time length of the ST-rPPG block as 10s. The time interval $\Delta t$ of each rPPG sample is half of the time length of the ST-rPPG block. We use AdamW optimizer \cite{loshchilov2018decoupled} to train our model with a learning rate of $10^{-5}$ for 30 epochs on one NVIDIA Tesla V100 GPU. For each training iteration, the inputs are two 10s clips from two different videos, respectively. During testing, we broke each test video into non-overlapping 30s clips and computed rPPG for each clip. We locate the highest peak in the PSD of an rPPG signal to calculate the HR. We use Neurokit2 \cite{makowski2021neurokit2} to calculate HRV metrics for the reported HRV results.

\subsubsection{Evaluation Metrics.} 
Following previous work \cite{li2014remote,niu2020video,yu2019remote}, we use mean absolute error (MAE), root mean squared error (RMSE), and Pearson correlation coefficient (R) to evaluate the accuracy of HR measurement. Following \cite{lu2021dual}, we also use standard deviation (STD), RMSE, and R to evaluate the accuracy of HRV features, including respiration frequency (RF), low-frequency power (LF) in normalized units (n.u.), high-frequency power (HF) in normalized units (n.u.), and the ratio of LF and HF power (LF/HF). For MAE, RMSE, and STD, smaller values mean lower errors, while for R, larger values close to one mean lower errors. Please check our supplementary material for more details about evaluation metrics.

\subsection{Intra-dataset Testing}

\subsubsection{HR Estimation.}

\begin{wrapfigure}{r}{0.35\linewidth}
\centering
\begin{minipage}[b]{\linewidth}
  \centering
  \centerline{\includegraphics[width=\linewidth]{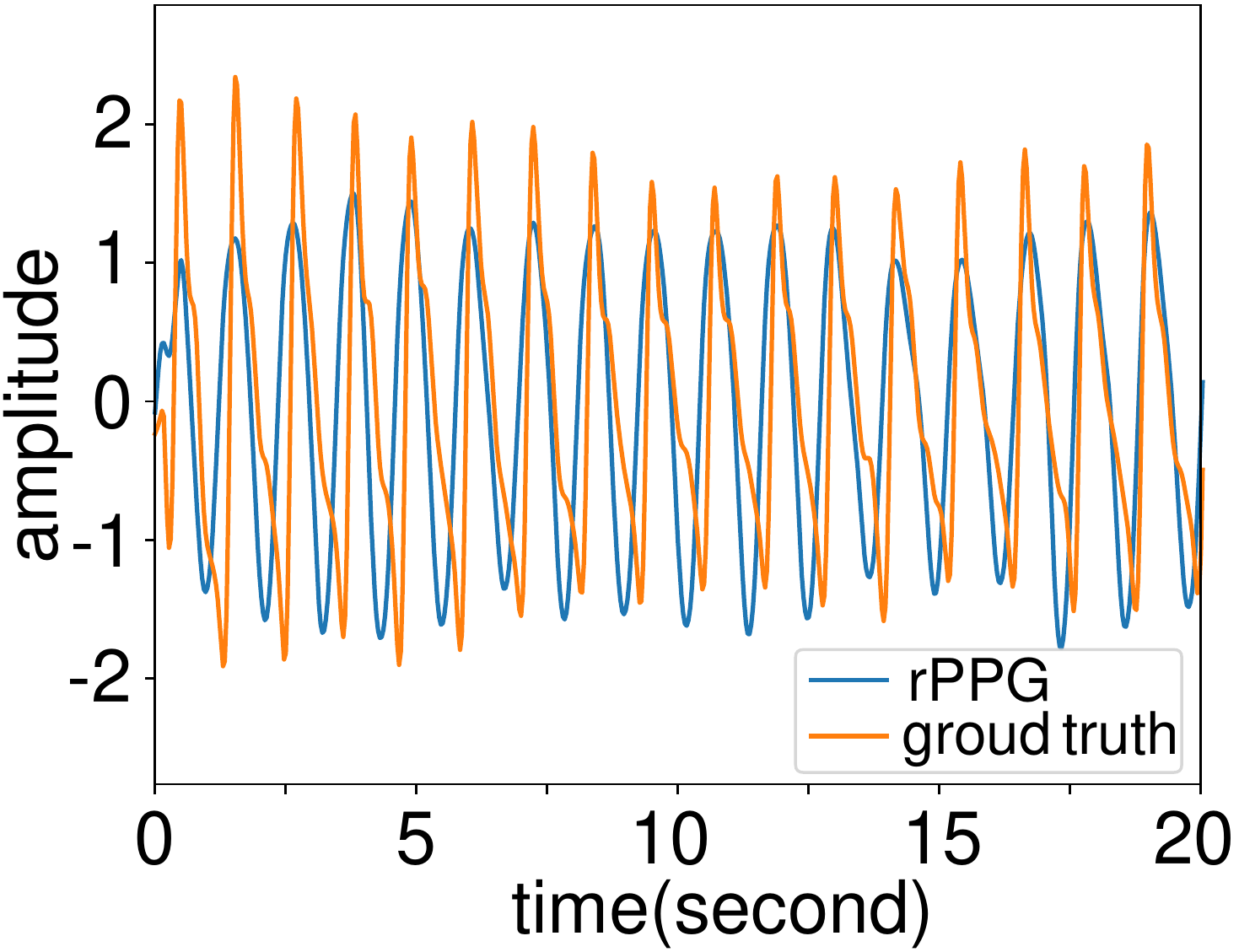}}
\end{minipage}
\caption{rPPG waveform and ground truth PPG signal.}
\label{fig:rppg_wave}
\end{wrapfigure}

We perform intra-dataset testing for HR estimation on PURE \cite{stricker2014non}, UBFC-rPPG \cite{bobbia2019unsupervised}, OBF \cite{li2018obf}, and MR-NIRP \cite{nowara2018sparseppg}. Table \ref{tb:intra} shows HR estimation results, including traditional methods, supervised methods, and unsupervised methods. The proposed Contrast-Phys largely surpasses the previous unsupervised baseline \cite{gideon2021way} and almost approaches the best supervised methods. The superior performance of Contrast-Phys is consistent on all four datasets, including the MR-NIRP \cite{nowara2018sparseppg} dataset, which contains NIR videos. The results indicate that the unsupervised Contrast-Phys can achieve reliable HR estimation on both RGB and NIR videos without requiring any ground truth physiological signals for training. Fig. \ref{fig:rppg_wave} also shows that the rPPG waveform from our unsupervised method is very similar to the ground truth PPG signal.

\begin{table}[htb!]
\caption{Intra-dataset HR results. The best results are in bold, and the second-best results are underlined.}
\fontsize{6}{8}\selectfont
\centering
{
\begin{tabular}{llcccccccccccc} 
\toprule
\multirow{2}{*}{\begin{tabular}[c]{@{}l@{}}\\Method\\ Types\end{tabular}}                                                                                                                   & \multirow{2}{*}{\begin{tabular}[c]{@{}l@{}}\\Methods\end{tabular}}  & \multicolumn{3}{c}{UBFC-rPPG}                                                                                                     & \multicolumn{3}{c}{PURE}                                                                                                          & \multicolumn{3}{c}{OBF}                                                                                                           & \multicolumn{3}{c}{MR-NIRP (NIR)}                                                                                                  \\ 
\cmidrule(lr){3-5}\cmidrule(lr){6-8}\cmidrule(lr){9-11}\cmidrule(lr){12-14}
& & \begin{tabular}[c]{@{}c@{}}MAE\\ (bpm)\end{tabular} & \begin{tabular}[c]{@{}c@{}}RMSE\\ (bpm)\end{tabular} & R                    & \begin{tabular}[c]{@{}c@{}}MAE\\ (bpm)\end{tabular} & \begin{tabular}[c]{@{}c@{}}RMSE\\ (bpm)\end{tabular} & R                    & \begin{tabular}[c]{@{}c@{}}MAE\\ (bpm)\end{tabular} & \begin{tabular}[c]{@{}c@{}}RMSE\\ (bpm)\end{tabular} & R                    & \begin{tabular}[c]{@{}c@{}}MAE\\ (bpm)\end{tabular} & \begin{tabular}[c]{@{}c@{}}RMSE\\ (bpm)\end{tabular} & R                     \\ 
\midrule
\multirow{5}{*}{\begin{tabular}[c]{@{}l@{}}Tradi-\\tional\end{tabular}}                                          & GREEN \cite{verkruysse2008remote} & 7.50 & 14.41 & 0.62 & - & - & - & - & 2.162 & 0.99 & - & - & - \\
& ICA \cite{poh2010advancements} & 5.17 & 11.76 & 0.65 & - & - & - & - & - & - & - & - & - \\
& CHROM \cite{de2013robust} & 2.37 & 4.91 & 0.89 & 2.07 & 9.92 & \bf 0.99 & - & 2.733 & 0.98 & - & - & - \\
& 2SR \cite{wang2015novel} & - & - & - & 2.44 & 3.06 & \underline{0.98} & - & - & - & - & - & -\\
& POS \cite{wang2016algorithmic} & 4.05 & 8.75 & 0.78 & - & - & - & - & 1.906 & 0.991 & - & - & - \\
 
\midrule
\multirow{8}{*}{\begin{tabular}[c]{@{}l@{}}Super-\\vised\end{tabular}} 

& CAN \cite{chen2018deepphys} & - & - & - & - & - & - & - & - & - & 7.78 & 16.8 & -0.03\\
& HR-CNN \cite{vspetlik2018visual} & - & - & - & 1.84 & 2.37 & \underline{0.98} & - & - & - & - & - & -\\
& SynRhythm \cite{niu2018synrhythm} & 5.59 & 6.82 & 0.72 & - & - & - & - & - & - & - & - & - \\
& PhysNet \cite{yu2019remoteBMVC} & - & - & - & 2.1 & 2.6 & \bf 0.99 & - & 1.812 & 0.992 & 3.07 & 7.55 & \underline{0.655}\\
& rPPGNet \cite{yu2019remote} & - & - & - & - & - & - & - & 1.8 & 0.992 & - & - & -\\
& CVD \cite{niu2020video} & - & - & - & - & - & - & - & \bf{1.26} & \bf{0.996} & - & - & -\\
& PulseGAN \cite{song2021pulsegan} & 1.19 & 2.10 & \underline{0.98} & - & - & - & - & - & - & - & - & - \\
& Dual-GAN \cite{lu2021dual} & \bf{0.44} & \bf{0.67} & \bf{0.99} & \bf{0.82} & \bf{1.31} & \bf{0.99} & - & - & - & - & - & - \\
& Nowara2021 \cite{nowara2021benefit} & - & - & - & - & - & - & - & - & - & \bf{2.34} & \bf{4.46} & \bf{0.85}\\
\midrule
 & Gideon2021 \cite{gideon2021way} & \begin{tabular}[c]{@{}l@{}}1.85 \end{tabular} & \begin{tabular}[c]{@{}l@{}}4.28 \end{tabular} & \begin{tabular}[c]{@{}l@{}}0.93 \end{tabular}   & 2.3 & 2.9 & \bf 0.99 & 2.83 & 7.88 & 0.825 & 4.75 & 9.14 & 0.61\\

 \multirow{-2}{*}{\begin{tabular}[c]{@{}l@{}}Unsup-\\ervised\end{tabular}} & \bf Ours & \begin{tabular}[c]{@{}l@{}} \underline{0.64} \end{tabular} & \begin{tabular}[c]{@{}l@{}} \underline{1.00} \end{tabular} & \begin{tabular}[c]{@{}l@{}} \bf{0.99} \end{tabular} & \underline{1.00} & \underline{1.40} & \bf 0.99 & \bf 0.51 & \underline{1.39} & \underline{0.994} & \underline{2.68} & \underline{4.77} & \bf 0.85 \\
\bottomrule
\end{tabular}}
\label{tb:intra}
\end{table}

\begin{table}[htb!]
\caption{HRV results on UBFC-rPPG. The best results are in bold, and the second-best results are underlined.}
\centering
\fontsize{6}{8}\selectfont
\begin{tabular}{llcccccccccccc} 
\toprule
\multirow{2}{*}{\begin{tabular}[c]{@{}l@{}} Method\\ Types\end{tabular}} & \multirow{2}{*}{\begin{tabular}[c]{@{}l@{}} Methods\end{tabular}}  & \multicolumn{3}{c}{LF (n.u.)}               & \multicolumn{3}{c}{HF (n.u.)} & \multicolumn{3}{c}{LF/HF} & \multicolumn{3}{c}{RF(Hz)} \\ 
\cmidrule(lr){3-5}\cmidrule(lr){6-8}\cmidrule(lr){9-11}\cmidrule(lr){12-14}&                          
& STD & RMSE & R & STD & RMSE & R & STD & RMSE & R & STD & RMSE & R
\\ 
\midrule
\multirow{3}{*}{\begin{tabular}[c]{@{}l@{}}Tradi-\\tional\end{tabular}}
& GREEN \cite{verkruysse2008remote} & 0.186 & 0.186 & 0.280 & 0.186 & 0.186 & 0.280 & 0.361 & 0.365 & 0.492 & 0.087 & 0.086 & 0.111 \\
& ICA \cite{poh2010advancements} & 0.243 & 0.240 & 0.159 & 0.243 & 0.240 & 0.159 & 0.655 & 0.645 & 0.226 & 0.086 & 0.089 & 0.102 \\
& POS \cite{wang2016algorithmic} & 0.171 & 0.169 & 0.479 & 0.171 & 0.169 & 0.479 & 0.405 & 0.399 & 0.518 & 0.109 & 0.107 & 0.087 \\
\midrule
\multirow{2}{*}{\begin{tabular}[c]{@{}l@{}}Super-\\vised\end{tabular}} 
& CVD \cite{niu2020video} & 0.053 & \underline{0.065} & 0.740 & 0.053 & \underline{0.065} & 0.740 & \underline{0.169} & \underline{0.168} & \underline{0.812} & \underline{0.017} & \underline{0.018} & 0.252\\
& Dual-GAN \cite{lu2021dual} & \bf{0.034} & \bf{0.035} & \bf{0.891} & \bf{0.034} & \bf{0.035} & \bf{0.891} & \bf{0.131} & \bf{0.136} & \bf{0.881} & \bf{0.010} & \bf{0.010} & \bf{0.395} \\
\midrule
& Gideon2021 \cite{gideon2021way} & 0.091 & 0.139 & 0.694 & 0.091 & 0.139 & 0.694 & 0.525 & 0.691 & 0.684 & 0.061 & 0.098 & 0.103 \\

\multirow{-2}{*}{\begin{tabular}[c]{@{}l@{}}Unsup-\\ervised\end{tabular}} & \bf Ours & \underline{0.050} & 0.098 & \underline{0.798} & \underline{0.050} & 0.098 & \underline{0.798} & 0.205 & 0.395 & 0.782 & 0.055 & 0.083 & \underline{0.347} \\
\bottomrule
\end{tabular}
\label{tb:HRV}
\end{table}

\subsubsection{HRV Estimation.}

We also perform intra-dataset testing for HRV on UBFC-rPPG \cite{bobbia2019unsupervised} as shown in Table \ref{tb:HRV}. HRV results need to locate each systolic peak, which requires high-quality rPPG signals. Our method significantly outperforms traditional methods and the previous unsupervised baseline \cite{gideon2021way} on HRV results. There is a marginal difference in HRV results between ours and the supervised methods \cite{niu2020video,lu2021dual}. The results indicate that Contrast-Phys can achieve high-quality rPPG signals with accurate systolic peaks to calculate HRV features, which makes it feasible to be used for emotion understanding \cite{yu2021facial,mcduff2014remote,sabour2021ubfc} and healthcare applications \cite{shi2019atrial,yan2018contact}.

\subsection{Cross-dataset Testing}

We conduct cross-dataset testing on MMSE-HR \cite{zhang2016multimodal} to test the generalization ability of our method. We train Contrast-Phys and Gideon2021 on UBFC-rPPG and test on MMSE-HR. We also show MMSE-HR cross-dataset results reported in the papers of some supervised methods \cite{niu2019rhythmnet,yu2019remoteBMVC,niu2020video,NEURIPS2020_e1228be4,nowara2021benefit} with different training sets. Table \ref{tb:cross} shows the cross-dataset test results. Our method still outperforms Gideon2021 \cite{gideon2021way} and is close to supervised methods. The cross-dataset testing results demonstrate that Contrast-Phys can be trained on one dataset without ground truth physiological signals, and then generalize well to a new dataset. 




\begin{table}[htb!]

\begin{minipage}[t][][b]{0.5\linewidth}
    \centering
    \caption{Cross-dataset HR Estimation on MMSE-HR. The best results are in bold, and the second-best results are underlined.}
    \fontsize{6}{8}\selectfont
    \begin{tabular}{llccc} 
    \toprule
    \begin{tabular}[c]{@{}l@{}} Method\\ Types\end{tabular} & \begin{tabular}[c]{@{}l@{}} Methods\end{tabular}& \begin{tabular}[c]{@{}l@{}} MAE\\ (bpm)\end{tabular} & \begin{tabular}[c]{@{}l@{}} RMSE\\ (bpm)\end{tabular} & R\\
    \midrule
    \multirow{3}{*}{\begin{tabular}[c]{@{}l@{}}Traditional\end{tabular}}
    & Li2014 \cite{li2014remote} & - & 19.95 & 0.38 \\
    & CHROM \cite{de2013robust} & - & 13.97 & 0.55 \\
    & SAMC \cite{tulyakov2016self} & - & 11.37 & 0.71 \\
    \midrule
    \multirow{5}{*}{\begin{tabular}[c]{@{}l@{}}Supervised\end{tabular}}
    & RhythmNet \cite{niu2019rhythmnet} & - & 7.33 & 0.78 \\
    & PhysNet \cite{yu2019remoteBMVC} & - & 13.25 & 0.44 \\
    & CVD \cite{niu2020video} & - & \underline{6.04} & 0.84 \\
    & TS-CAN \cite{NEURIPS2020_e1228be4} & 3.41 & 7.82 & 0.84 \\
    & Nowara2021 \cite{nowara2021benefit} & \bf{2.27} & \bf{4.90} & \bf{0.94} \\
    \midrule
    & Gideon2021 \cite{gideon2021way} & 4.10 & 11.55 & 0.70 \\
     \multirow{-2}{*}{\begin{tabular}[c]{@{}l@{}}Unsupervised\end{tabular}} & \bf Ours & \underline{2.43} & 7.34 & \underline{0.86} \\
    \bottomrule
    \end{tabular}
    \label{tb:cross}
  \end{minipage}%
  \hfill
  \begin{minipage}[t][][b]{0.45\linewidth}
    \centering
    \includegraphics[width=\linewidth]{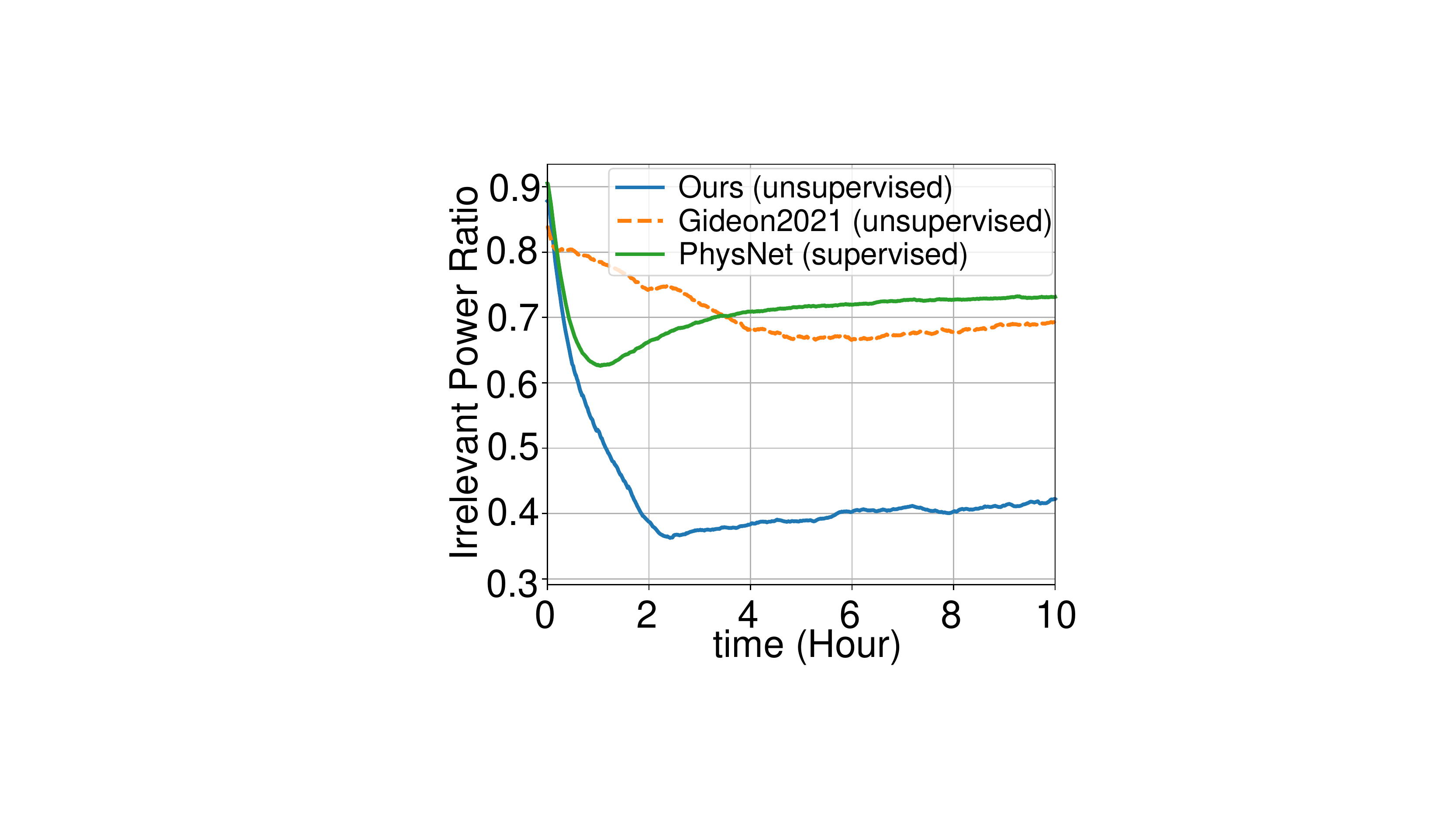}
    \captionof{figure}{irrelevant power ratio change during training time for unsupervised and supervised methods}
    \label{fig:speed}
  \end{minipage}
\end{table}

\subsection{Running Speed}

We test the running speed of the proposed Contrast-Phys and compare it with Gideon2021 \cite{gideon2021way}. During training, the speed of our method is \textbf{802.45} frames per second (fps), while that of Gideon2021 is \textbf{387.87} fps, which is about half of our method's speed. The large difference is due to different method designs. For Gideon2021, one input video has to be fed into the model twice, i.e., firstly as the original video and then as a temporal resampled video for a second time, which causes double computation. For our method, one video is fed into the model once, which can substantially decrease computational cost compared with Gideon2021.

Furthermore, we compare the convergence speed of the two unsupervised methods and one supervised method (PhysNet \cite{yu2019remoteBMVC}) using the metric of irrelevant power ratio (IPR). IPR is used in \cite{gideon2021way} to evaluate the signal quality during training, and lower IPR means higher signal quality. (more details about IPR are in the supplementary materials.) Fig. \ref{fig:speed} shows IPR with respect to time during training on the OBF dataset. Our method converges to the lowest IPR point for about 2.5 hours, while Gideon2021 \cite{gideon2021way} converges to the lowest point for about 5 hours. In addition, the lowest IPR for our method is about 0.36 while that for Gideon2021 \cite{gideon2021way} is about 0.66 which is higher than ours. The above evidence demonstrates that our method converges faster to a lower IPR than Gideon2021 \cite{gideon2021way}. In addition, our method also achieves lower IPR than the supervised method (PhysNet \cite{yu2019remoteBMVC})

\subsection{Saliency Maps}

\begin{figure}[hbt!]
\centering
\begin{minipage}[b]{\linewidth}
  \centering
  \centerline{\includegraphics[width=\linewidth]{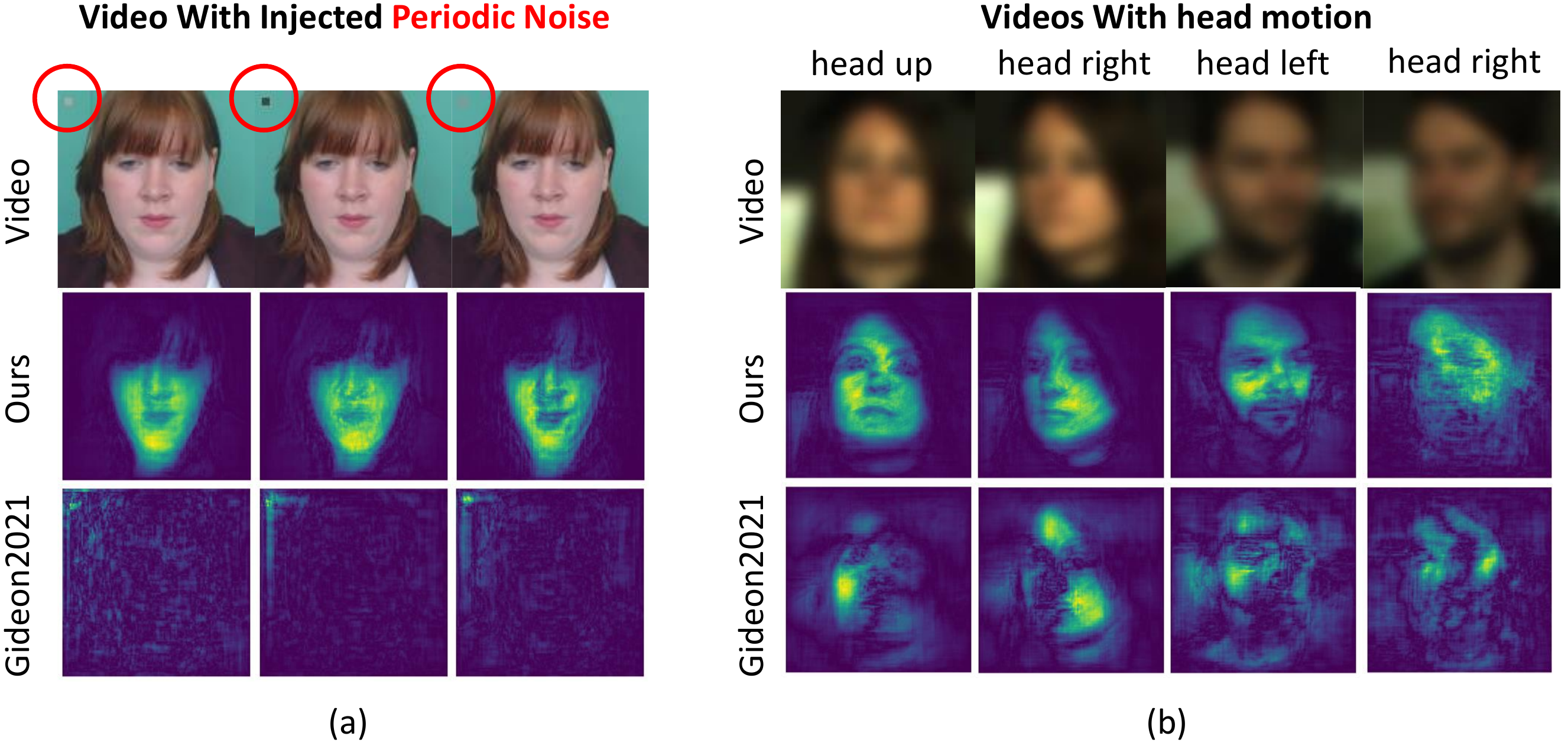}}
\end{minipage}
\caption{Saliency maps for our method and Gideon2021 \cite{gideon2021way} (a) We add a random flashing block with the HR range between 40-250 bpm in the top left corner to all UBFC-rPPG videos. The models for our method and Gideon2021 \cite{gideon2021way} are trained on these videos with the noise. Our saliency maps have a high response on facial regions, while Gideon2021 focuses on this injected periodic noise. Table \ref{tb:noise} also shows that our method is robust to the noise. (b) We choose the head motion moments in PURE dataset (Video frames shown here are blurred due to privacy issues.) and show the saliency maps for our method and Gideon2021 \cite{gideon2021way}.}
\label{fig:saliency}
\end{figure}

We calculate saliency maps to illustrate the interpretability of our method. The saliency maps are obtained using a gradient-based method proposed in \cite{simonyan2013deep}. We fix the weights of the trained model and get the gradient of Pearson correlation with respect to the input video (More details are in the supplementary materials.). Saliency maps can highlight spatial regions from which the model estimates the rPPG signals, so the saliency map of a good rPPG model should have a large response on skin regions, as demonstrated in \cite{yu2019remoteBMVC,yu2019remote,gideon2021way,chen2018deepphys,nowara2021benefit}. 

\begin{wraptable}{r}{0.45\linewidth}
    \centering
    \caption{HR results trained on UBFC-rPPG with/without injected periodic noise shown in Fig. \ref{fig:saliency}(a)}
    \scriptsize
    \begin{tabular}{lcccc} 
    \toprule
    \begin{tabular}[c]{@{}l@{}} Methods\end{tabular} & \begin{tabular}[c]{@{}l@{}} Injected \\ Periodic \\ Noise\end{tabular}& \begin{tabular}[c]{@{}l@{}} MAE\\ (bpm)\end{tabular} & \begin{tabular}[c]{@{}l@{}} RMSE\\ (bpm)\end{tabular} & R\\
    \midrule
    \multirow{2}{*}{\begin{tabular}[c]{@{}l@{}}Gideon2021 \cite{gideon2021way}\end{tabular}}
    & w/o & 1.85 & 4.28 & 0.939 \\
    & w/ & 22.47 & 25.41 & 0.244 \\
    \midrule
    \multirow{2}{*}{\begin{tabular}[c]{@{}l@{}}Ours\end{tabular}}
    & w/o & 0.64 & 1.00 & 0.995 \\
    & w/ & 0.74 & 1.34 & 0.991 \\
    \bottomrule
    \end{tabular}
    \label{tb:noise}
\end{wraptable} 

Fig. \ref{fig:saliency} shows saliency maps under two cases, 1) periodic noise is manually injected, and 2) head motion is involved. When a periodic noise patch is injected to the left-upper corner of videos, our method is not distracted by the noise and still focuses on skin areas, while Gideon2021 is completely distracted by the noise block. We also calculate the two methods' performance on UBFC-rPPG videos with the injected noise, and the results are listed in Table \ref{tb:noise}. The results are consistent with the saliency map analysis, that the periodic noise does not impact Contrast-Phys, but fails Gideon2021 completely. Our method is robust to the noise because the noise only exists in one region, which violates rPPG spatial similarity. Fig. \ref{fig:saliency}(b) shows the saliency maps when head motion is involved. The saliency maps for our method focus and activate most skin areas, while saliency maps for Gideon2021 \cite{gideon2021way} show messy patterns and only partially cover facial areas during head motions.

\subsection{Sensitivity Analysis}
We perform sensitivity analysis on two variables: 1) the spatial length $S$ of the ST-rPPG block, and 2) the temporal length $T$ of the ST-rPPG block.

Table \ref{tb:ablation}(a) shows the HR results on UBFC-rPPG, when the ST-rPPG spatial resolution is set in four levels of 1x1, 2x2, 4x4 or 8x8. Note that $1 \times 1$ means that rPPG spatial similarity is not used. The results of $1 \times 1$ are worse than other results with spatial information, which means rPPG spatial similarity improves performance. In addition, $2 \times 2$ is enough to perform well since larger resolutions do not significantly improve the HR estimation. Although 8x8 or 4x4 provides more rPPG samples, it has a smaller receptive field and thus produces noisier rPPG samples than a 2x2 block.

Table \ref{tb:ablation}(b) shows the HR results on UBFC-rPPG, when the ST-rPPG temporal length is set on three levels of 5s, 10s and 30s. The results indicate that 10s is the best choice. A shorter time length (5s) causes coarse PSD estimation, while a long time length (30s) causes a slight violation of short-term signal condition in rPPG temporal similarity. Therefore, in these two cases (5s and 30s), the performance is lower than that of 10s.

\begin{table}[htb!]
\centering
\caption{Sensitivity Analysis: (a) HR results on UBFC-rPPG with different ST-rPPG block spatial resolutions. (b) HR results on UBFC-rPPG with different ST-rPPG block time lengths. (The best results are in bold.)}
  \begin{minipage}[t][][b]{0.45\linewidth}
    \centering
    \centerline{(a)}
    \fontsize{8}{8}\selectfont
    \begin{tabular}{lccc} 
    \toprule
     \begin{tabular}[c]{@{}l@{}} Spatial \\ Resolution \end{tabular}& \begin{tabular}[c]{@{}l@{}} MAE\\ (bpm)\end{tabular} & \begin{tabular}[c]{@{}l@{}} RMSE\\ (bpm)\end{tabular} & R\\
    \midrule
     $1 \times 1$ & 3.14 & 4.06 & 0.963 \\
     $\bf{2 \times 2}$ & \bf 0.64 & \bf 1.00 & \bf 0.995 \\
     $4 \times 4$ & 0.55 & 1.06 & 0.994 \\
     $8 \times 8$ & 0.60 & 1.09 & 0.993 \\
    \bottomrule
    \end{tabular}
  \end{minipage}
  \begin{minipage}[t][][b]{0.45\linewidth}
    \centering
    \centerline{(b)}
    \fontsize{8}{9.6}\selectfont
    \begin{tabular}{lccc} 
    \toprule
     \begin{tabular}[c]{@{}l@{}} Time \\ Length\end{tabular}& \begin{tabular}[c]{@{}l@{}} MAE\\ (bpm)\end{tabular} & \begin{tabular}[c]{@{}l@{}} RMSE\\ (bpm)\end{tabular} & R\\
    \midrule
     5s & 0.68 & 1.36 & 0.990 \\
     \bf 10s & \bf 0.64 & \bf 1.00 & \bf 0.995 \\
     30s & 1.97 & 3.58 & 0.942 \\
    \bottomrule
    \end{tabular}
  \end{minipage}
  \label{tb:ablation}
\end{table}

\section{Conclusion}

We propose Contrast-Phys which can be trained without ground truth physiological signals and achieve accurate rPPG measurement. Our method is based on four observations about rPPG and utilizes spatiotemporal contrast to enable unsupervised learning. Contrast-Phys significantly outperforms the previous unsupervised baseline \cite{gideon2021way} and is on par with the state-of-the-art supervised rPPG methods. In the future work, we would like to combine the supervised and the proposed unsupervised rPPG methods to further improve performance.

\subsubsection{Acknowledgment.} The study was supported by Academy of Finland (Project 323287 and 345948) and the Finnish Work Environment Fund (Project 200414). The authors also acknowledge CSC-IT Center for Science, Finland, for providing computational resources.

%
%
\newpage
\bibliographystyle{splncs04}
\bibliography{egbib}
\end{document}